\pdfoutput=1

\documentclass[11pt]{article}

\usepackage{EMNLP2022}

\usepackage{times}
\usepackage{latexsym}
\usepackage{graphicx}
\usepackage{bbm}
\usepackage{booktabs}
\usepackage{subfigure}
\usepackage{amsmath}
\usepackage[T1]{fontenc}

\usepackage[utf8]{inputenc}

\usepackage{microtype}

\usepackage{inconsolata}
\usepackage{multirow}
\usepackage[ruled,linesnumbered]{algorithm2e}
\usepackage{amsfonts} 

\title{Neural Machine Translation with Contrastive Translation Memories}

\newcommand\icst{$^1$}
\newcommand\tencent{$^4$}
\newcommand\data{$^2$}
\newcommand\sdu{$^3$}
\newcommand\ruc{$^7$}
\newcommand\xhs{$^5$}
\newcommand\bigai{$^6$}

\author{
Xin Cheng \icst$^{,}$\data, 
Shen Gao \sdu, 
Lemao Liu \tencent, 
Dongyan Zhao \icst$^{,}$\data$^{,}$\xhs$^{,}$\bigai\thanks{\;\;Corresponding author.},  
Rui Yan \ruc $^*$ \\
\icst~Wangxuan Institute of Computer Technology, Peking University\\
\data~Center for Data Science, Peking University \quad \tencent~Tencent AI Lab\\ 
\sdu~School of Computer Science and Technology, Shandong University\\
\xhs~State Key Laboratory of Media Convergence Production Technology and Systems \\
\bigai Beijing Institute of General Artificial Intelligence (BIGAI)\\
\ruc~Gaoling School of Artificial Intelligence, Renmin University of China\\
{\tt chengxin1998@stu.pku.edu.cn}\quad
{\tt shengao@sdu.edu.cn} \\
{\tt zhaody@pku.edu.cn} \quad
{\tt redmondliu@tencent.com} \quad
{\tt ruiyan@ruc.edu.cn}\\
}

\begin{document}
\maketitle
\begin{abstract}
Retrieval-augmented Neural Machine Translation models have been successful in many translation scenarios. Different from previous works that make use of mutually similar but redundant translation memories~(TMs), we propose a new retrieval-augmented NMT to model contrastively retrieved translation memories that are holistically similar to the source sentence while individually contrastive to each other providing maximal information gains in three phases. First, in TM retrieval phase, we adopt a contrastive retrieval algorithm to avoid redundancy and uninformativeness of similar translation pieces. Second, in memory encoding stage, given a set of TMs we propose a novel Hierarchical Group Attention module to gather both local context of each TM and global context of the whole TM set. Finally, in training phase, a Multi-TM contrastive learning objective is introduced to learn salient feature of each TM with respect to target sentence. Experimental results show that our framework obtains  improvements over strong baselines on the benchmark datasets.
\end{abstract}
\section{Introduction}
 
Translation memory (TM) is basically a database of segmented and paired source and target texts that translators can access in order to re-use previous translations while translating new texts~\cite{christensen2010translation}. For human translators, such similar translation pieces can lead to higher productivity and consistency~\citep{yamada2011effect}. For machine translation, early works mainly 
contributes to employ TM for statistical machine translation (SMT) systems ~\citep{simard2009phrase,utiyama2011searching,liu-etal-2012-locally,liu2019unified}. Recently, as neural machine translation (NMT) model~\cite{seq2seq,transformer} has achieved impressive performance in many translation tasks, there is also an emerging interest  ~\citep{gu} in retrieval-augmented NMT model.

\begin{figure}
    \centering
    \includegraphics[width=0.45\textwidth]{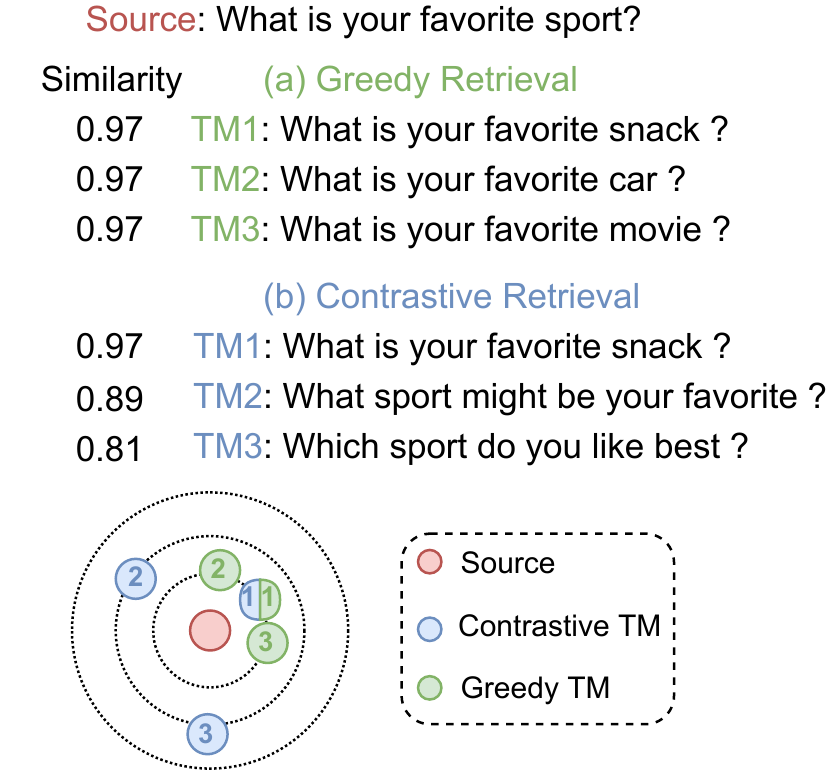}
    \caption{An example of \textit{Greedy Retrieval }and \textit{Contrastive Retrieval}. The similarity score is computed by edit distance detailed in Section \ref{section:contrastive retrieval}. And the target side of TM is omitted for brevity.}
    \label{fig:intro}
\end{figure}

The key idea of retrieval-augmented NMT mainly includes two steps: a retrieval metric is used to retrieve similar translation pairs (i.e., TM), and the TM is then integrated into an NMT model. 
In the first step, a standard retrieval method greedily chooses the most similar translation memory one by one solely based on similarity with the source sentence (namely \textit{Greedy Retrieval}). This method would inevitably retrieve translation memories that are mutually similar but redundant and uninformative as shown in Figure~\ref{fig:intro}. Intuitively, it is promising to retrieve a diverse translation memory which would offer maximal coverage of the source sentence and provide useful cues from different aspects. Unfortunately, empirical experiments in \citet{gu} show that a diverse translation memory only leads to negligible improvements. As a result, greedy retrieval  is adopted in almost all later studies~\cite{gated,xia,boosting,qiuxiang,cai,knn_mt}.

This paper aims to ask an important question \textit{whether diverse translation memories are beneficial for retrieval-augmented NMT}. To this end, we propose a powerful retrieval-augmented NMT model called Contrastive Memory Model which takes into account diversity in translation memory from three ways. 
Specifically, 
(1)~during TM retrieval, inspired by Maximal Marginal Relevance~(MMR)~\cite{mmr}, we introduce a conceptually simple while empirically useful retrieval method called \textit{Contrastive Retrieval} to find informative translation memories. The core is to retrieve a cluster of translation memories that are similar to the source sentence while contrastive to each other keeping inner-cluster uniformity in the latent semantic space, as shown in Figure~\ref{fig:intro}.
(2)~In TM encoding, given multiple translation memories, the local and global information should both be captured by the translation model. Separately encoding~\citep{gu,qiuxiang,cai} or treating them as a long sequence~\citep{boosting} would inevitably lose such hierarchical structure information. 
Thus, to facilitate the direct communication between different translation memories for local information and gather the global context via message passing mechanism, we propose a Hierarchical Group Attention (HGA) module to encode the diverse memories.
(3)~In the model training phase, to learn salient and distinct features of each TM with respect to target sentence, we devise a novel Multi-TM Contrastive Learning objective (MTCL), which further contributes to a uniformly distributed translation memory cluster by forcing representation of every translation memory to approach the sentence to be translated while keep away from each other. 

To verify the effectiveness of our framework, we conduct extensive experiments on four benchmark datasets, and observe substantial improvement over strong baselines, proving that diverse translation memories is indeed useful to NMT. Our main contributions are:

\begin{itemize}
    \item We answer an important question about retrieval-augmented NMT, i.e., is diverse translation memory beneficial for retrieval-augmented NMT?
    \item We propose a diverse-TM-aware framework to improve a retrieval-augmented NMT system from three ways including TM retrieval, TM encoding and model training.
    \item We conduct extensive experiments on four translation directions, observing substantial performance gains over strong baselines with greedy retrieval.
\end{itemize}

\section{Related Work}
\noindent\textbf{TM-augmented NMT} Augmenting neural machine translation model with translation memories is an important line of work to boost the performance of the NMT model with non-parametric method. \citet{feng2017memory} stores memories of infrequently encountered words and utilizes them to assist the neural model. \citet{gu} uses an external memory network and a gating mechanism to incorporate TM. \citet{gated} uses an extra GRU-based memory encoder to provide additional information to the decoder. \citet{xia} adopts a compact graph representation of TM and perform additional attention mechanisms over the graph when decoding. ~\citet{fuzzy} and ~\citet{boosting} directly concatenate TM with source sentence using cross-lingual vocabulary. \citet{zhang} augments the model with an additional bonus given to outputs that contain the collected translation pieces. There is also a line of work that trains a parametric retrieval model and a translation model jointly~\citep{cai} and achieves impressive results.
Recently, with rapid growth of computational power, a more fine grained token level translation memories are use in \citet{knn_mt}. This approach gives the decoder direct access to billions of examples at test time, achieving state-of-the-art result even without further training.\\

\noindent\textbf{Contrastive Learning} The key of contrastive learning~\cite{hadsell2006dimensionality,mikolov2013distributed} is to learn effective representation by pulling semantically close neighbors together and pushing apart non-neighbors. 
\citet{chen2020simple} and ~\citet{moco} show that contrastive learning can boost the performance of self-supervised and semi-supervised learning in computer vision tasks. In natural language processing, Word2Vec~\cite{mikolov2013distributed} uses noise-contrastive estimation to learn better word representation. \citet{simcse} adopts contrastive learning with a simple token level dropout to greatly advance the state-of-the-art sentence embeddings. \citet{simcls} uses contrastive loss to post-rank generated summaries and achieves promising results in benchmark datasets. \citet{lee2020contrastive} and ~\citet{mrasp2} also use contrastive learning in translation tasks and observe consistent improvements.

\section{Proposed Framework}
\label{section:nmt}

\paragraph{Preliminary}
Assuming we are given a source sentence
\(x = \{\mathtt{x}_1,...,\mathtt{x}_{s}\}\) and its corresponding target sentence
\(y= \{\mathtt{y}_1,...,\mathtt{y}_{t}\}\) where $s,t$ are their respective length. For a TM-augmented NMT model, a set of similar translation pairs \(M=\{(x^m,y^m)\}_{m=1}^{|M|}\) are retrieved based on certain criterion $\mathbb{C}$ and NMT models the conditional probability of target sentence \(y\) conditioned on both source sentence $x$ and translation memories $M$ in a left-to-right manner:
\begin{equation}
\label{equa:nmt}
\begin{aligned}
    P(Y=y|X=x)=\prod_{t=1}^{|T|}P(\mathtt{y}_t|&\mathtt{y}_0,...\mathtt{y}_{t-1};x;M)
\end{aligned}
\end{equation}

\begin{figure*}
    \centering
    \includegraphics[width=0.95\textwidth,height=0.65\textwidth]{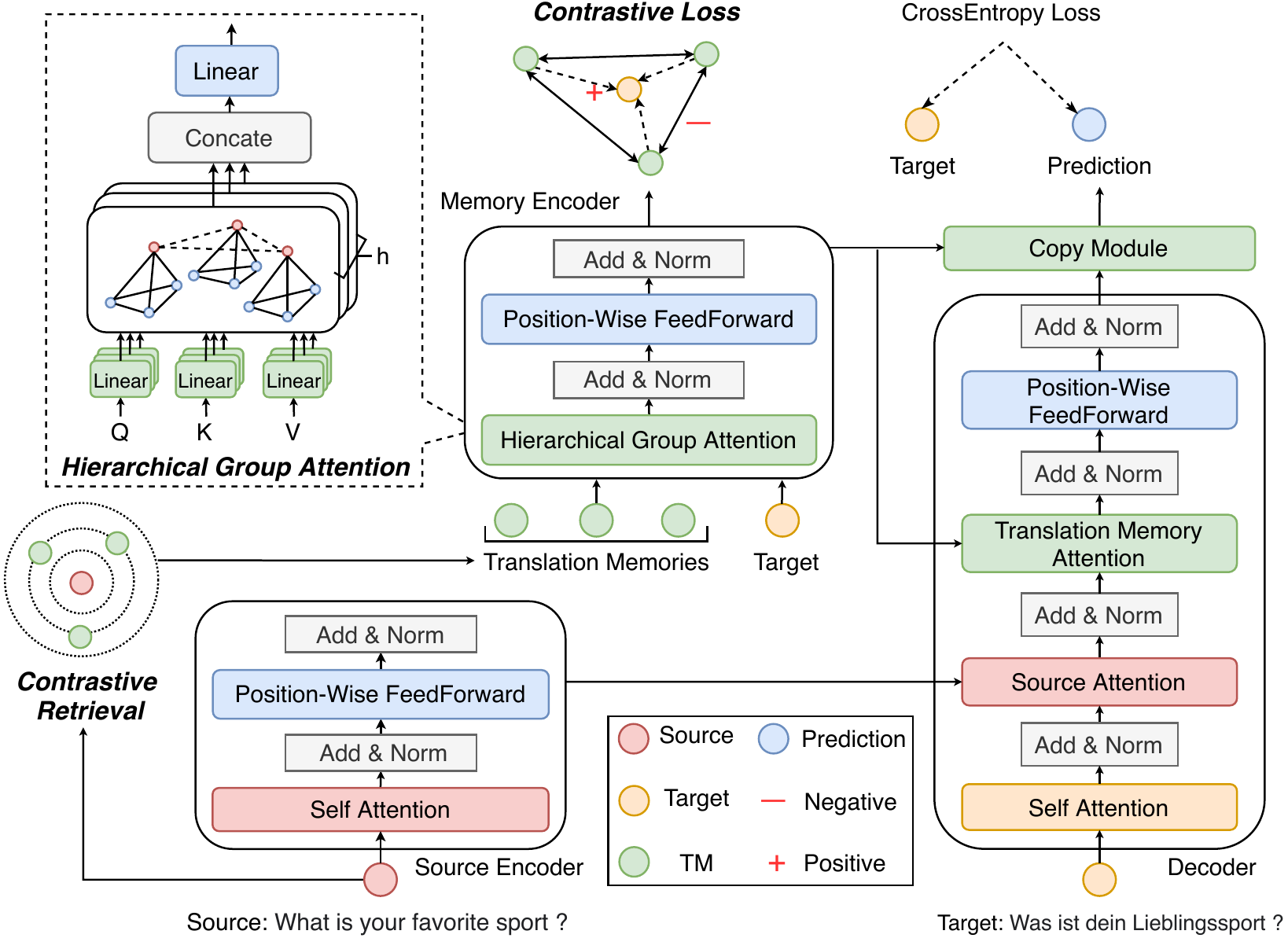}
    \caption{Overview of our framework: (1) \textbf{\textit{Contrastive Retrieval}} (2) Source Encoder; (3) Memory Encoder with \textbf{\textit{Hierarchical Group Attention}} module (we only show three translation memories for brevity); (4) Decoder; (5) \textbf{\textit{Contrastive Learning}}. }
    \label{fig:model}
\end{figure*}

\paragraph{Overview}
Given a source sentence $x$ and informative translation memories $M$ , the translation model defines the conditional probability similar to the
Equation~\ref{equa:nmt}. At the high level, our framework, as shown in Figure~\ref{fig:model}, consists of \textit{contrastive retrieval}, which searches a diverse translation memory,  \textit{source encoder} which
transforms source sentence $x$ into dense vector
representations $z^x$, \textit{memory encoder} with hierarchical
group attention module to jointly encode $|M|$ translation memories into a series dense representation $z^m$ and \textit{decoder} which attends to both $z^x$ and $z^m$ and generates target sentence $y$ in an auto-regressive fashion, 
and \textit{contrastive learning} which effectively trains the memory encoder as well as source encoder and decoder. 
\textit{Among all these five modules,  contrastive memory (\S 3.1), memory encoder (\S 3.3) and {contrastive learing} (\S 3.5) are key in our framework compared with existing work of TM-augmented NMT.} 

\subsection{Contrastive Retrieval}
\label{section:contrastive retrieval}
In this stage, following previous work~\citep{gu} we first employ an off-the-shelf full-text search engine, Apache Lucene, to get a preliminary translation memory set $K=\{(x^k,y^k)\}_{k=1}^{|K|}~(|K|\gg|M|)$ for every source sentence. Notice that both source sentence $x$ and translation memory set $K$ are from training set $\mathcal{D}=\{(x^n,y^n)\}_{n=1}^N$, which means we do not introduce any extra data during training. Then to be directly comparable with previous works~\cite{gu,qiuxiang} as discussed in Section \ref{sec:results} and considering the core of our method is similarity function-agnostic as detailed below, we adopt a sentence-level similarity function:
\begin{equation}
\text{sim}(x,x') = 1-\frac{D_{\text{edit}}(x,x')}{\text{max}(|x|,|x'|)}  
\end{equation}
\noindent where \(D_{\text{edit}}\) is the edit distance between two sentences and \(|x|\) is the length of $x$. Specifically, we would select $|M|$ translation memories incrementally and in every step we do not only measure the similarity between current translation memory and the source sentence but also take into consideration the edit distance with those already retrieved ones balanced by a hyperparameter $\alpha$~(namely contrastive factor). Different from MMR~\cite{mmr}, we treat retrieved translation memories as a whole and take the average similarity score as a penalty term:
\begin{equation}
\mathop{\arg\max}\limits_{x^i\in K\backslash M}\,[\text{sim}(x,x^i) - \frac{\alpha}{|M|}\sum\limits_{x^j\in M}\text{sim}(x^i,x^j)]    
\end{equation}
\noindent where $M$ is the post-ranked translation memory set. Finally, for every source sentence $x$, by ignoring the source side of $M$ due to information redundancy we have translation memories \(M=\{y^m\}_{m=1}^{|M|}\).

\subsection{Source Encoder}
For a source sentence $x=\{\mathtt{x}_1,...,\mathtt{x}_s\}$, our \textit{source encoder} is built upon the standard Transformer~\cite{transformer} architecture composed of a token embedding layer, a sinusoidal positional embedding Layer and stacked transformer encoder layers. Specifically we prepend a \textit{<bos>} token to source sentence and get the dense vector representation $z^x$ as follows:

\begin{equation}
z^x=\text{SrcEnc}(\text{<bos>},\mathtt{x}_1,...\mathtt{x}_{s})
\end{equation}

\subsection{Memory Encoder}
\label{sec:mem_enc}
Given a set of translation memories, the local context of each TM and the global context of the whole TM set should be captured by the model to fully utilize this hierarchical structure information. Separately encoding~\citep{gu,cai} or treating them as a long sequence~\citep{boosting} would inevitably mask the  model with this kind of local and global schema.
In this section, to facilitate the direct communication between different translation memories for local information and gather the global context via message passing mechanism, we propose a Hierarchical Group Attention (HGA) module. Formally, given a cluster of translation memories $M=\{y^m\}_{m=1}^{|M|}$, where each $y^m=\{y_0^m,...,y_{n_m}^k\}$ is composed of $n_m$ tokens, for each $y^m$ we would like to create a fully connected graph $G^m=(V^m,E^m)$  where $V^m$ is the token set.
To facilitate inter-memory communication, we also create a super node $v_{*}^m$ by connecting it with all other nodes~(namely trivial node) in that graph and then connect all super nodes together contributing to information flow among different translation memories in a hierarchical way as shown in Figure \ref{fig:model}. Then we adopt multi-head self attention 
mechanism~\citep{transformer} as message passing operator~\cite{gilmer2017neural}. For every node $v_i^m$ in the graph, their hidden state in time step $t+1$ is updated by the hidden states of its neighbours $\phi(v_i^m)$ in time step $t$:
\begin{equation}
    v_i^m|_{t+1} = \text{SelfAttn}(\phi(v_i^m|_t),v_i^m|_t)
\end{equation}
\noindent To be computationally efficient, we use mask mechanism to block communication between nodes in different graphs. For each trivial node $v_i^m$ in $G^m$, they update their hidden states by attending to all trivial nodes as well as super node~$v^m_{*}$. For $v^m_{*}$, it does not only exchange information within the graph $G^m$, but also communicate with all other super nodes $\{v_{*}^i\}_{i=1}^{|M|}$. 
To stabilize training, we also add residual connection and feed-forward Layer after HGA module. After stacking multiple layers, we get dense representation of translation memories:
\begin{equation}
z^m=\text{MemEnc}(\text{Concate}\{y^m\}_{m=1}^{|M|})
\end{equation}

\noindent where $|m|$ is the total length of $|M|$ translation memories and $z^m \in \mathbb{R}^{|m|\times d}$.

\subsection{Fusing TM in Decoding}
To better incorporate the information from both source sentence
\(z^x\) and translation memories \(z^m\), we introduce a multi-reference
decoder architecture. For a target sentence
$y$ , we get a hidden representation \(h=\{h_1,...,h_{t}\}\) after token embedding layer and masked self-attention layer, then we use a cross attention layer to fuse information
from source sentence:
\begin{equation}
    \hat{h} = \text{CrossAttn}(\text{Add\&Norm}(h),z^x,z^x)
\end{equation}

\noindent Then for translation memories, we employ another cross attention layer:
\begin{equation}
    \overline{h} = \text{CrossAttn}(\text{Add\&Norm}(\hat{h}),z^m,z^m)
\end{equation}

\noindent After stacking multiple decoder layers, to further exploit translation
memories, we apply a copy module ~\citep{pointer_generator,gu2016incorporating} using the attention score
from the second cross attention layer in the last sub-layer of decoder
as a probability of directly copying the corresponding token from the translation memory. Formally,
with \(t-1\) previous generated tokens and hidden state~\(h_t\), the
decoder computes \(t\)-th token probability as:
\begin{equation}
  p(y_t|\cdot) = (1-p_{\text{copy}})p_{v}(y_t) + p_{\text{copy}}\sum_{i=1}^{|z^m|}\alpha_i\mathbbm{1}_{z_i^{m}=y_t}  
\end{equation}

\noindent where \(p_{\text{copy}} = \sigma (MLP(h_t,y_{t-1},\alpha\otimes z^m))\) , \(\alpha\)
is the attention score,
$\otimes$ is a Hadamard product and \(\mathbbm{1}\) is the
indicator function.

\subsection{Multi-TM Contrastive Learning}

The key of contrastive learning is to learn effective representation by pulling semantically close neighbors together and pushing apart non-neighbors~\cite{hadsell2006dimensionality,mikolov2013distributed}. As indicated in~\citep{lee2020contrastive}, simply choosing in-batch negatives would yield meaningless negative examples that are already well-discriminated in the embedding space and would even cause performance degradation in translation task~\citep{lee2020contrastive}, which also holds true in our preliminary experiments. So how to devise effective contrastive learning objective for a translation model with a cluster of translation memories to learn salient features with respect to the target sentence remains unexplored and challenging.

In this work, to make every translation memory learn distinct and useful feature representations with respect to current target sentence, we propose a novel Multi-TM Contrastive Learning (MTCL) objective
which do not simply treat in-batch samples as negative but instead keep
aligned with the principle of our \textit{contrastive retrieval}, making every translation memory approach the ground truth translation while pushing apart from each other. Formally, given a source sentence $x$, its corresponding target sentence $y$ and translation memories \(M = \{y^m\}_{m=1}^{|M|}\). The goal of MTCL is to minimize the
following loss:
\begin{equation}
\mathcal{L}_{\text{MTCL}} =- \sum_{y^i\in M}log\frac{e^{\text{sim}(y^i,y)/\tau}}{\sum_{y^j \in M}e^{\text{sim}(y^j,y)/\tau}}    
\end{equation}
\noindent where $\text{sim}(y^i,y)$ is the cosine similarity between the representation of target sentence $y$ and translation memory $y^i$ given by \textit{memory encoder} and $\tau$ is a temperature hyperparameter which controls the difficulties of distinguishing between positive and negative samples~\citep{mrasp2}. Notice that the representation of each translation memory is the super node $v_*^m$ given by HGA module in Section \ref{sec:mem_enc}, which communicates with both intra-memory and inter-memory nodes.
Intuitively, by maximizing the softmax term $e^{\text{sim}(y^i,y)/\tau}$, the contrastive loss would force the representation of each translation memories to approach the ground truth while push apart from each other, delivering a uniformly distributed representation around the target sentence in latent semantic space.
In MTCL, all negative samples are not from in-batch data but are different translation memories for one source sentence, which make up of non-trivial negative samples and help the model to learn the subtle difference between multiple translation memories.

During the training phase, the model can be optimized by jointly minimizing the MTCL loss and Cross Entropy loss as shown:
\begin{equation}
\mathcal{L} = \mathcal{L}_{\text{CE}}+\lambda\mathcal{L}_{\text{MTCL}}
\end{equation}
where \(\lambda\) is a balancing coefficient to measure the importance
of different objectives in a multi-task learning scenario~\cite{sener2018multi}.\\
\section{Experimental Setup}

\subsection{Dataset and Evaluation}
We use the JRC-Acquis~\citep{steinberger2006jrc} corpus to evaluate our model. This corpus is a collection of parallel legislative text of European union Law applicable in the EU member states. Highly related and well structured data make this corpus an ideal test bed to evaluate the proposed TM-augmented translation model. Following previous work, we use the  same split of train/dev/test set as in \citep{gu,xia,cai,boosting,qiuxiang}. 
For evaluation, we use \textit{SacreBLEU}.

\subsection{Implementation Details}
Our model is named Contrastive Memory Model~(CMM). 
To implement CMM, we use transformer as building block of our model. Specifically, we adopt the base configuration and the default optimization configuration as in \citet{transformer}. We use joint BPE encoding~\citep{bpe} with vocab size 35000. We also adopt label smoothing as 0.1 in all experiments. The number of tokens in every batch is 10000, which includes both source sentence and translation memories. The memory size and contrastive factor is set to be 5 and 0.7 across all translation directions. The contrastive temperature~$\tau$ is $\{0.1,0.08,0.05,0.15\}$ for Es$\rightarrow$En, En$\rightarrow$Es, De$\rightarrow$En and En$\rightarrow$De directions. The balancing factor $\lambda$ is set to be 1~\footnote{Code and data is available at \url{https://github.com/Hannibal046/NMT_with_contrastive_memories}}.

\subsection{Baselines}

CMM is compared with the following baselines:
\noindent $\bullet$ \citet{transformer}: this is the original implementation of base transformer.

\noindent $\bullet$ \citet{gu}: this is a pioneer work of integrating translation memories into NMT system using an external memories networks to separately encode every translation memory

\noindent $\bullet$ \citet{boosting}: this paper augments source sentence with concatenation of TM and euqip the model with different language embedding~(FM$^{+}$).

\noindent $\bullet$ \citet{xia}: this work uses a compact graph to encode translation memories and is also based on transformer architecture.

\noindent $\bullet$ \citet{zhang}: this work equips a NMT model with translation pieces and extra bonus given to outputs that contain the collected translation pieces.

\noindent $\bullet$ \citet{cai}: this model first retrieves translation memories by source side similarity and adopts a dual encoder architecture.

\noindent $\bullet$ \citet{qiuxiang}: this model incorporates one most similar translation memory with proposed example layer.

In addition, considering that \citet{gu} is based on Memory Network and RNN architecture, to be fairly compared with transformer based model, we re-implement two more direct baselines (i.e., {\bf BaseGreedy} and {\bf T-Ada}) on top of Transformer with the same configuration as our CMM. Specifically, in both baselines the original Memory Network is replaced by a transformer encoder sharing weights with source encoder. BaseGreedy employs greedy retrieval and it does not take diversity of TM into account. In contrast, T-Ada adopts adaptive retrieval, which finds the translation memories via maximizing the token coverage of source sentence,  and it promotes the diversity in retrieved memory to some extent as CMM. 

\begin{table}
\centering
\resizebox{\columnwidth}{!}{
\begin{tabular}{c|c|c|c|c}
	     \bottomrule
	     \multicolumn{2}{c|}{}&		CMM	&	T-Ada & BaseGreedy \\
	     \hline
 	     \hline
	     \multicolumn{2}{l|}{Avg. TM Size}	&		5		&	5.68 & 5\\ 
	     \multicolumn{2}{l|}{Avg. Coverage}	&		84.01\%	&	92.11 \% & 81.13\%\\
	     \multicolumn{2}{l|}{Avg. Similarity}&		0.89	&	0.84 & 0.91\\
  	     \hline
		 \multicolumn{2}{l|}{Training Latency}&		1.21x	&	1.25x  & 1.21x\\
	     \multicolumn{2}{l|}{Inference Latency}&		1.44x	&	1.56x & 1.44x\\
	     \hline
	     \multirow{4}[2]{*}{\rotatebox{90}{BLEU}} 
	     &Es$\rightarrow$En&		\textbf{67.76}$\dagger$		& 67.08 & 66.84\\
	     &En$\rightarrow$ES&		\textbf{64.04}$\dagger$		& 63.56 & 63.18\\
	     &De$\rightarrow$En&		\textbf{64.33}$\dagger$		& 63.81 & 63.84\\
	     &En$\rightarrow$De&		\textbf{58.69}$\dagger$	& 57.28 & 57.02\\
	     \bottomrule
	
\end{tabular}
}
\caption{Comparison between CMM, T-Ada and BaseGreedy. The TM Size, Coverage and Similarity is averaged among four translation directions. Coverage means the token level coverage of all translation memories with respect to source sentence. Similarity score is calculated as described in Section~\ref{section:contrastive retrieval}. $\dagger$ means CMM is significantly better than baselines with \textit{p-value} < 0.01.}
\label{table:adaptive}
\end{table}

\section{Experiment Results}
\label{sec:results}

\begin{table*}
\centering
\begin{tabular}{c || c c|c c|c c|c c}
     \bottomrule
     \multirow{2}{4em}{System} &\multicolumn{2}{c}{Es$\rightarrow$En} & \multicolumn{2}{c}{En$\rightarrow$Es} &\multicolumn{2}{c}{De$\rightarrow$En} &\multicolumn{2}{c}{En$\rightarrow$De} \\
     \cline{2-9}
     &Dev&Test&Dev&Test &Dev&Test &Dev&Test \\
     \hline
     \hline
     \citealp{transformer}$\dagger$ & 64.08 & 64.63 & 62.02 & 61.80 & 60.18 & 60.16 & 54.65 & 55.07 \\
     \citealp{gu}  & 57.62 & 57.27 &  60.28  & 59.34  & 55.63 & 55.33  & 49.26  & 48.80 \\
     \citealp{zhang} & 63.97 & 64.30 & 61.50 & 61.56 & 60.10 & 60.26 & 55.54 & 55.14 \\
     \citealp{boosting}* & 66.44 & 65.90 &   -   &  -  & -  & -  & -  & - \\
     \citealp{xia} & 66.37 & 66.21 & 62.50 & 62.76 & 61.85 & 61.72 & 57.43 & 56.88 \\ 
     \citealp{qiuxiang}(@s) & 67.23 & 67.26 &   -   &  -  & -  & -  & -  & - \\
     \citealp{cai}(\#2) & 66.98 & 66.48 & 63.04 & 62.76 & 63.62 & 63.85 & 57.88 & 57.53 \\
     CMM & \textbf{67.48} & \textbf{67.76} & \textbf{63.84} & \textbf{64.04} & \textbf{64.22}& \textbf{64.33} &\textbf{58.94} & \textbf{58.69} \\
     \bottomrule
\end{tabular}
\caption{BLEU points on four translation directions of JRC-Acquis dataset. $\dagger$ denotes that the model is implemented by ourselves. @s means the model is trained under standard training criterion and * means results are from \citet{qiuxiang}. \#2 is the second model proposed in~\citet{cai} using source retrieval.}
\label{table:main}
\end{table*}

\begin{table*}
\centering
\resizebox{2\columnwidth}{!}{
\begin{tabular}{c|c|c|c| c c|c c| cc|cc}
     \bottomrule
     \multirow{2}{4em}{System} &\multirow{2}{5em}{Model Size} & \multirow{2}{4em}{Training} & \multirow{2}{4em}{Inference}& \multicolumn{2}{c}{Es$\rightarrow$En} &\multicolumn{2}{c}{En$\rightarrow$Es} &\multicolumn{2}{c}{De$\rightarrow$En} &\multicolumn{2}{c}{En$\rightarrow$De} \\
    \cline{5-12}
     &	&	&	&	Dev	&	Test&	Dev	&	Test&	Dev	&	Test&	Dev	&	Test\\
     \hline
     \hline
     T-Para    &   101M &   2.76x   &   1.36x   &   \textbf{67.73}   &   67.42   &   \textbf{64.18}  &   63.86 & \textbf{64.48} & \textbf{64.62} & 58.77 &58.42 \\
     CMM   &   68M  &   1.21x   &   1.44x   &  67.48 & \textbf{67.76} & 63.84 & \textbf{64.04} & 64.22& 64.33 &\textbf{58.94} & \textbf{58.69} \\
     \bottomrule
\end{tabular}
}
\caption{Translation quality and running efficiency compared with the strong model T-Para. }
\label{table:cai}
\end{table*}

\subsection{Main results}
\paragraph{Is diverse translation memory helpful?}
We make a comparison with the direct baseline T-Ada because both the proposed CMM and T-Ada promote the diversity in translation memory. As shown in Table~\ref{table:adaptive}, T-Ada yields modest gains (about +0.2 BLEU points on average) over BaseGreedy on four translation tasks, which is in line with the results in \citet{gu} on the RNN architecture. 
We conjecture that it is because \textit{Adaptive Retrieval} only partially maximize the word coverage while neglecting the overall semantics of the whole sentence thus injecting undesirable noise into the retrieval phase.
In contrast, the proposed CMM takes both token-level coverage and sentence-level similarity into consideration
and consistently outperforms T-Ada, gaining about 0.5-1.4 BLEU points on four tasks in translation quality with smaller TM size and lower latency in both training and inference phase. This fact shows the following findings: 1) NMT augmented with diverse translation memory can yield consistent improvements in translation quality; 2) how to model and learn the diverse translation memory is important in addition to promoting diversity in translation memory. Because of the potential problem of high BLEU test~\citep{callison-burch-etal-2006-evaluating}, we conduct another two experiments. First, We use metrics other than BLEU to evaluate our high BLEU systems. We compare our model CMM and BaseGreedy in JRC/EsEn dataset. We use both model-free and model-based metrics as shown in 
Table~\ref{table:metrics}. A clear patent here is that our higher-BLEU model CMM outperforms BaseGreedy model in all these metrics. Second, we disengage our model from high BLEU range by picking the hard sentences from the test set of JRC/EsEn according to the sentence-level BLEU for a vanilla Transformer model. The evaluation results for top-25\%, top-50\%, top-75\% hardest subsets are shown in Table~\ref{table:picking}. We can see that the proposed CMM still outperforms baselines on the top-25\% subset whose BLEU is in the range of 30s.

\begin{table}
\centering
\resizebox{\columnwidth}{!}{
\begin{tabular}{c|c|c|c|c|c}
	     \bottomrule
	     & BLEU	& Chrf & TER & BertScore & BartScore \\
	     \hline
 	     \hline
	     BaseGreedy	&	66.84	& 78.45 & 25.39 &   0.9686 & 0.1209 \\ 
	     CMM & \bf 67.76 & \bf 79.01 & \bf 24.43 & \bf 0.9698 & \bf 0.1329 \\
	     \bottomrule
	
\end{tabular}
}
\caption{Evaluation results with different metrics.}
\label{table:metrics}
\end{table}

\begin{table}[ht]
\centering
\resizebox{\columnwidth}{!}{
\begin{tabular}{c|c|c|c|c}
	     \bottomrule
	     & top-25\%	& top-50\% & top-75\% & ALL  \\
	     \hline
 	     \hline
	     \citet{transformer} &	29.17	& 43.48 & 56.07 & 64.63 \\ 
	     BaseGreedy &	34.17	& 48.77 & 59.94 & 66.84 \\ 
	     CMM & \bf 35.38 & \bf 49.37 & \bf 60.53 & \bf 67.76\\
	     \bottomrule
	
\end{tabular}
}
\caption{Evaluation results in terms of BLEU in different difficulty range.}
\label{table:picking}
\end{table}

\paragraph{Comparing with other baselines}
Since our CMM involves the heuristic metric (i.e., TF-IDF and normalized edit distance) for retrieval, we first compare our methods with other works using the same retrieval metric.  
The result is presented in Table \ref{table:main}. As can be seen, our method yields consistent better results than all other baseline models across four tasks in terms of BLEU. 
Substantial improvement by an average 3.31 BLEU points and up to 4.29 in En$\rightarrow$De direction compared with transformer baseline model demonstrates the effectiveness of incorporating translation memories into NMT model. 
In comparison with previous works either using greedy retrieval~\citep{gu,zhang,xia,cai}, which introduces redundant and uninformative translation memories, or using top1 similar translation memory~\citep{boosting,qiuxiang}, which causes omission of potentially useful cues, our framework equipped with contrastive translation memories can deliver consistent improvement in both development set and test set among four translation directions.

Unlike the above work, there is also another line of work that retrieve translation memory with a learnable metric. \citet{cai} proposes a powerful framework (namely T-Para) which jointly trains the retrieval metric and translation model in an end-to-end fashion, leading to state-of-the-art performance in translation quality. We also compare our method with this strong model and result is shown in Table~\ref{table:cai}. Notice that our model gives comparable results with T-Para, which is actually remarkable considering that our model has much smaller model size and training latency. In particular, our work about contrastive translation memory is orthogonal to \citet{cai} and it is promising to apply our idea into their framework, which remains a future work.

\begin{table}
\small
  \centering
    \begin{tabular}{l|l|cc|cc}
    \toprule
    \multicolumn{2}{c|}{\multirow{2}[2]{*}{}}& \multicolumn{2}{c|}{En-De} & \multicolumn{2}{c}{En-Es} \\
    \multicolumn{2}{c|}{} & \multicolumn{1}{c}{$\rightarrow$} & \multicolumn{1}{c|}{$\leftarrow$} & \multicolumn{1}{c}{$\rightarrow$} & \multicolumn{1}{c}{$\leftarrow$} \\

    \midrule
    \multirow{5}[2]{*}{\rotatebox{90}{Dev}} 
          & T w/o MTCL & \multicolumn{1}{c}{58.55} & 64.14 & \multicolumn{1}{c}{63.26} & \multicolumn{1}{c}{67.30} \\
          & T w/o HGA & \multicolumn{1}{c}{58.06} & 63.85 & 62.74 & \multicolumn{1}{c}{67.28} \\
          & T-Greedy & \multicolumn{1}{c}{58.01} & 63.72 & 63.10 & \multicolumn{1}{c}{66.98} \\
          & T-MMR & \multicolumn{1}{c}{58.20} & 64.10 & 62.66 & \multicolumn{1}{c}{67.25} \\
       	  & CMM & \multicolumn{1}{c}{\textbf{58.94}} & \textbf{64.22} & \multicolumn{1}{c}{\textbf{63.84}} & \multicolumn{1}{c}{\textbf{67.48}}\\
    \midrule
    \multirow{5}[2]{*}{\rotatebox{90}{Test}} 
		  & T w/o MTCL & \multicolumn{1}{c}{58.37} & 64.29 & \multicolumn{1}{c}{63.92} & \multicolumn{1}{c}{67.49} \\
          & T w/o HGA & \multicolumn{1}{c}{58.06} & 64.19 & 62.74 & \multicolumn{1}{c}{67.28} \\
          & T-Greedy & \multicolumn{1}{c}{57.66} & 63.57 & 63.28 & \multicolumn{1}{c}{67.16} \\
          & T-MMR & \multicolumn{1}{c}{57.95} & 64.27 & 63.10 & \multicolumn{1}{c}{67.15} \\
          & CMM & \multicolumn{1}{c}{\textbf{58.69}} & \textbf{64.33} & \multicolumn{1}{c}{\textbf{64.04}} & \multicolumn{1}{c}{\textbf{67.76}}\\

    \bottomrule
    \end{tabular}%
     \caption{Ablation study in four translation tasks with respect to each key component in our framework.}
  \label{table:ablation}%
\end{table}

\subsection{Analysis}
\paragraph{Ablation Study}
We also implement several variants of our framework: 
(1) T w/o MTCL: this model uses the same model configuration as CMM but without MTCL loss.
(2) T w/o HGA: in this setting, $|M|$ translation memories are concatenated together as a long sequence without Hierarchical Group Attention module.
(3) T-Greedy: this model replaces the \textit{Contrastive Retrieval} in CMM by \textit{Greedy Retrieval}.
(4) T-MMR: this model replaces the \textit{Contrastive Retrieval} in CMM by Maximal Marginal Relevance~\cite{mmr} while the setting of translation model keeps the same as CMM.
The result is shown in Table \ref{table:ablation} and we have the following observations. 
Simply replacing \textit{Contrastive Retrieval} by \textit{Greedy Retrieval} or MMR while keeping the setting of translation model unchanged yields worse results than our model which demonstrates that the informative translation memories serve as key ingredient in a TM-augmented NMT model. 
Interestingly, direct removal of HGA module while maintaining MTCL objective (i.e.,\ T w/o HGA) gives consistent worse results in four translation directions. We suspect that a pull-and-push game brought by contrastive learning causes performance  degradation without modeling the fine-grained interaction among multiple translation memories. Combining HGA and MTCL, which facilitates communication between different translation memories and helps the model to learn the subtle difference between them, performs better than all other baseline models revealing the fact that properly designed contrastive learning objective and HGA module is complementary to each other.\\

\begin{figure}
  \centering
  \includegraphics[width=0.24\textwidth,height=0.2\textwidth]{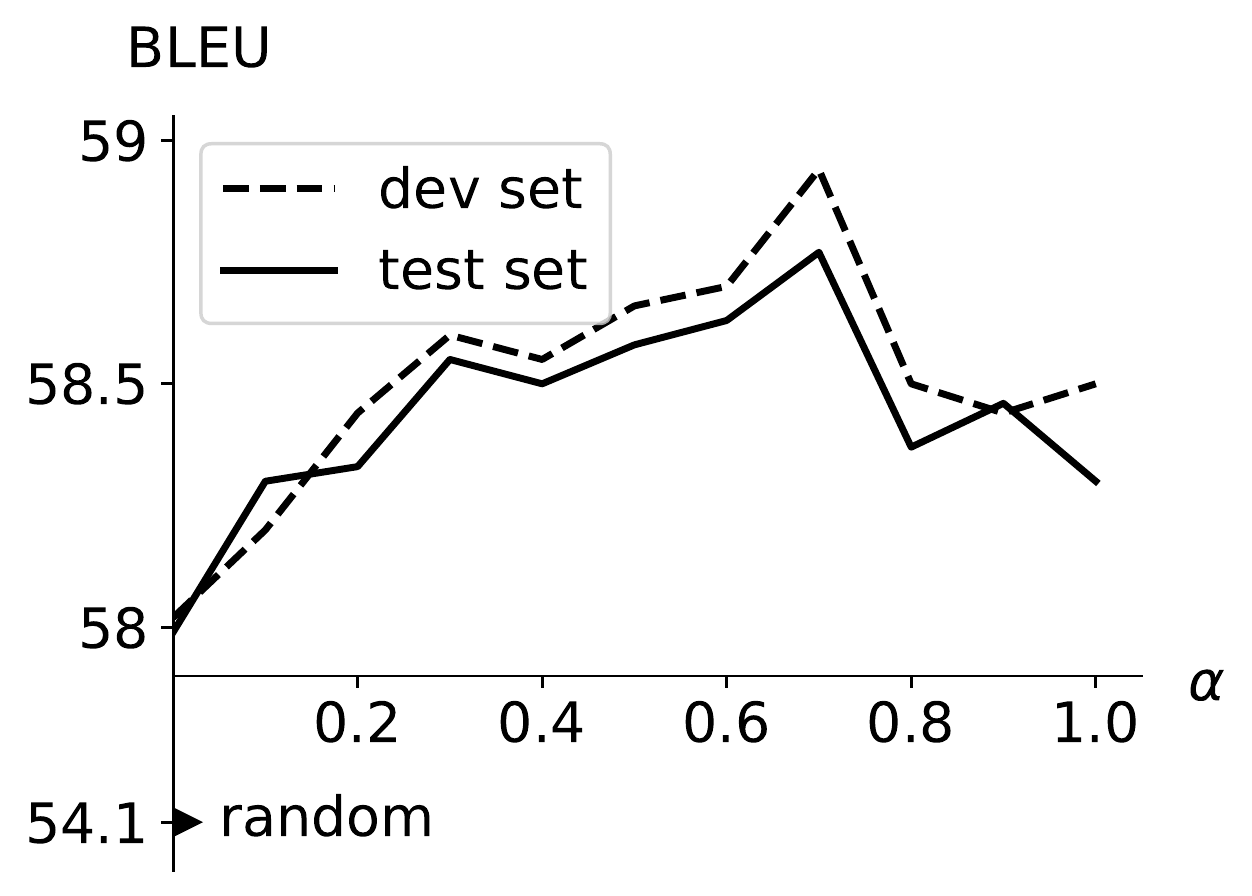}
  \includegraphics[width=0.22\textwidth,height=0.2\textwidth]{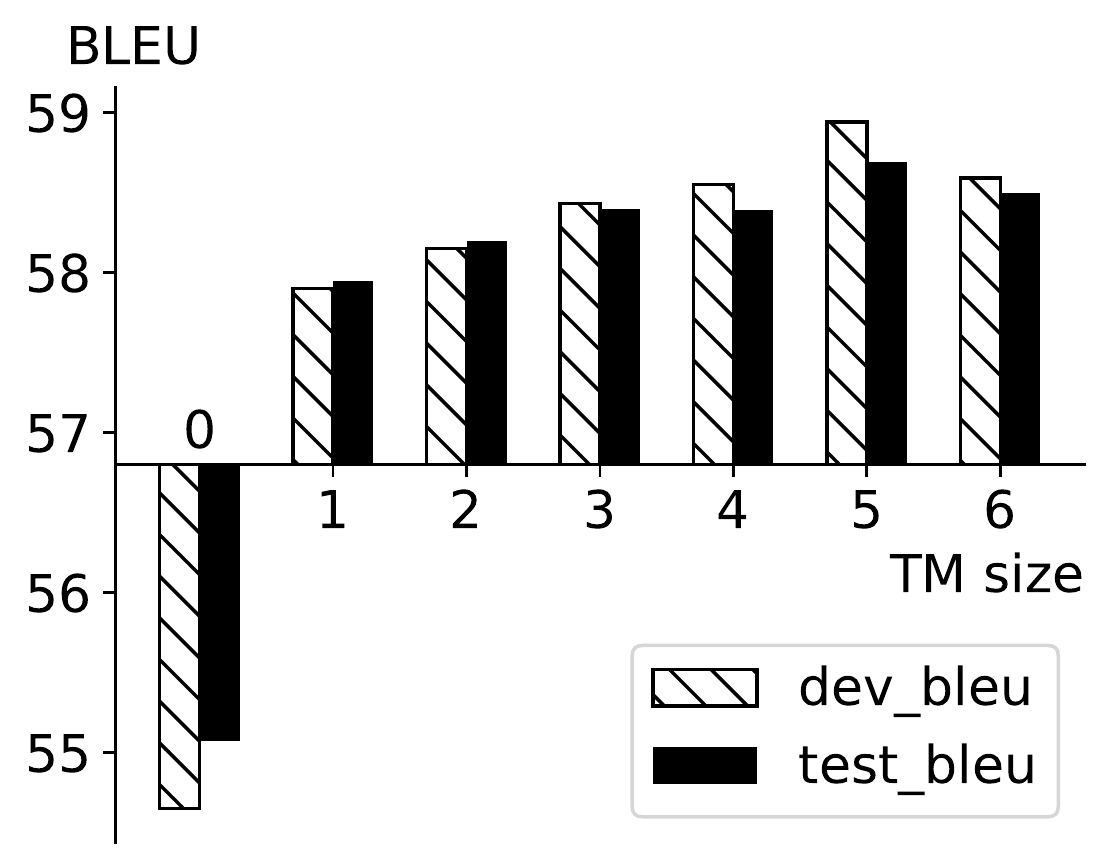}
  \caption{Effect of contrastive factor and TM size.}
  \label{fig:tm_size}
\end{figure}

\begin{figure}
  \centering
  \includegraphics[width=0.23\textwidth,height=0.23\textwidth]{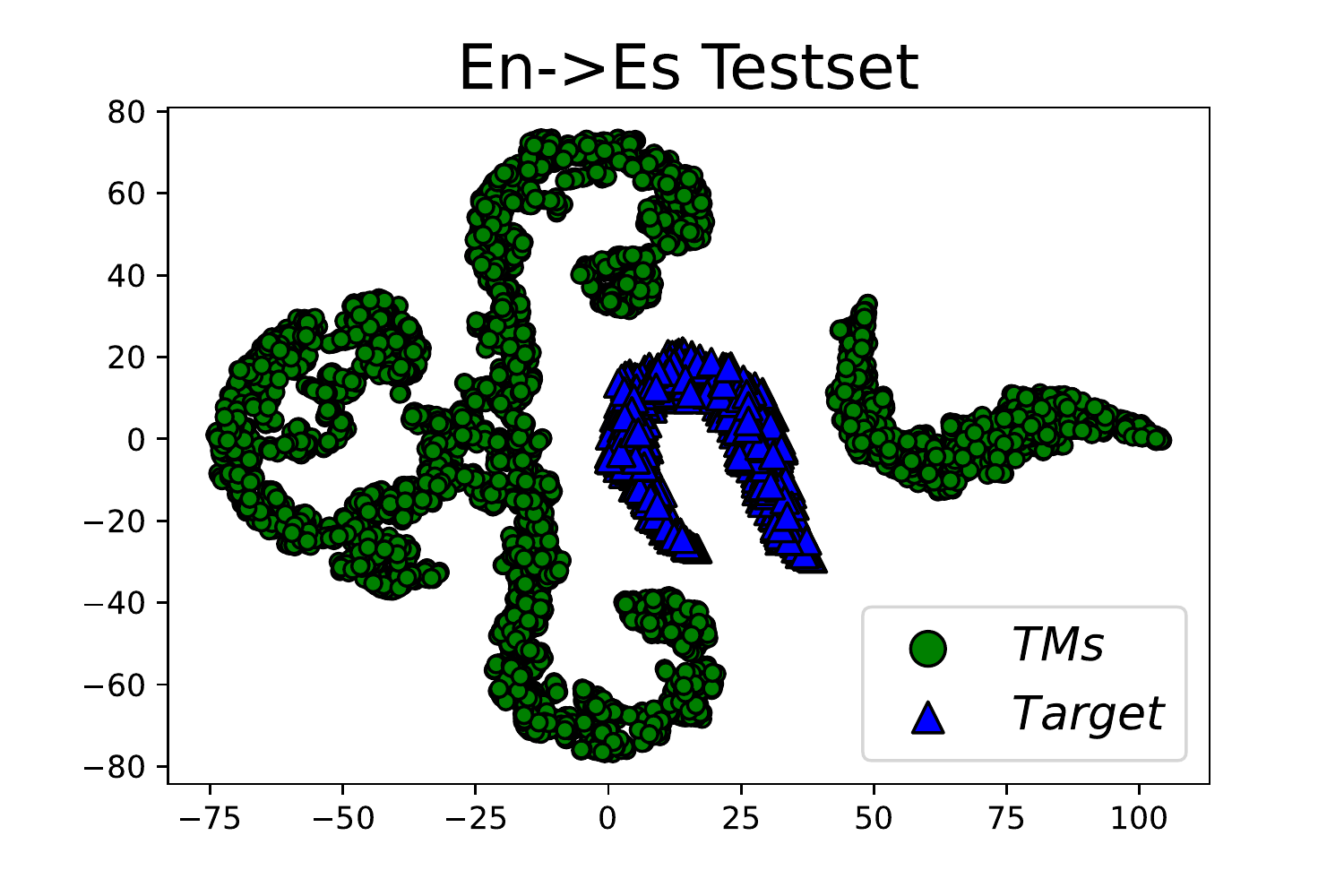}
  \includegraphics[width=0.23\textwidth,height=0.23\textwidth]{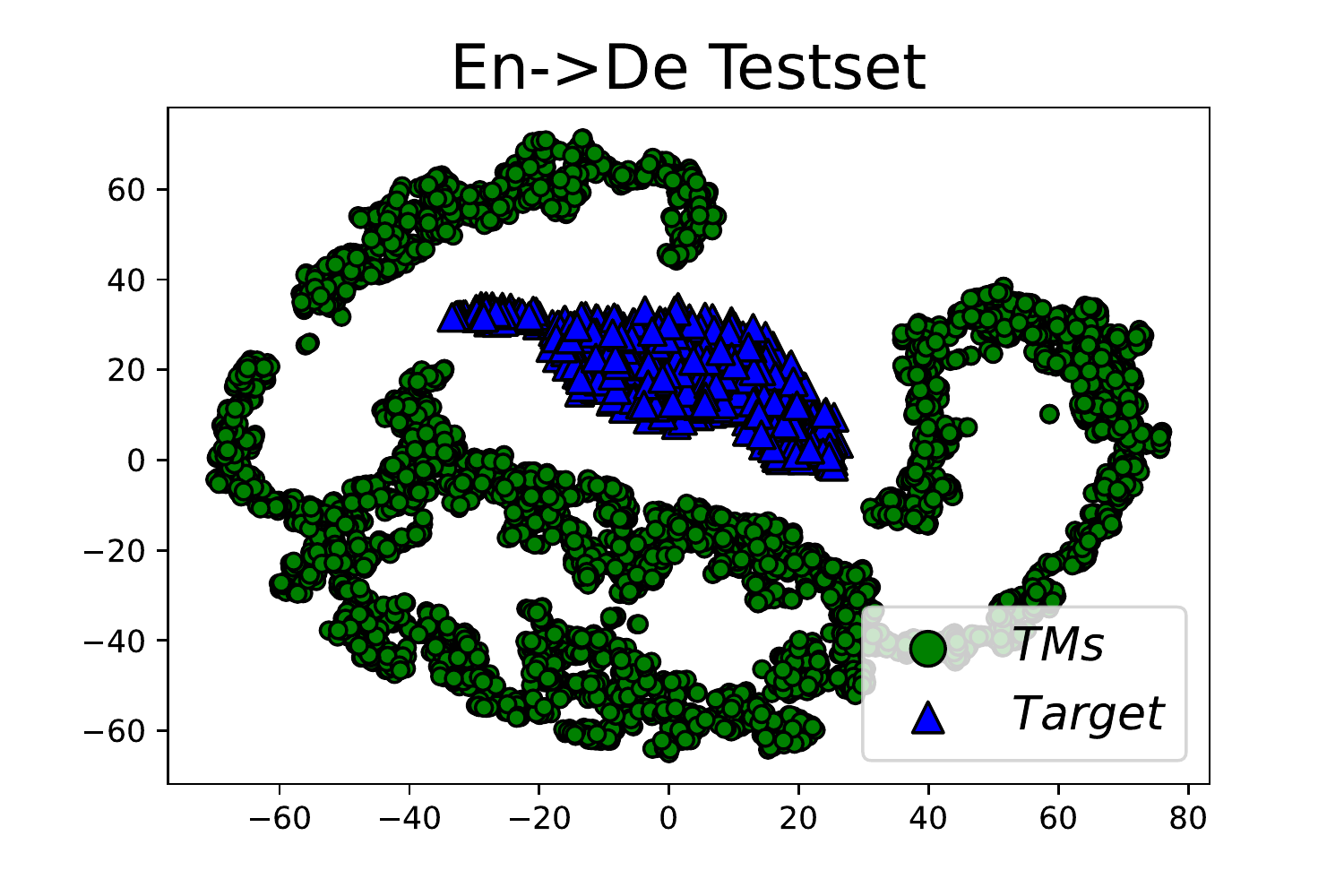}
  \caption{Visualization of Translation Memories and Target sentence in EnEs and EnDe testset by t-SNE.}
  \label{fig:visualization}
\end{figure}

\paragraph{Memory Size and Contrastive Factor}
To verify the effectiveness of fusing multiple contrastive translation memories, we choose En$\rightarrow$De dataset and make the following experiments in both TM retrieval and TM fusion stage:
In retrieval stage, we explore the contrastive factor $\alpha$ which is supposed to decide the degree of currently retrieved translation memory contrasting to those already retrieved. A larger $\alpha$ indicates that the retrieved translation memories are less similar to the source sentence while more contrastive to each other. And in fusion stage, the size $|M|$ of translation memories is considered. 
The effect of different $\alpha$ is shown in Figure \ref{fig:tm_size}. The \textit{random} point is the result of a NMT model with $|M|$ randomly retrieved translation memories and it even underperforms a non-TM translation model~\citep{transformer} shown in Table \ref{table:main}. We assume it is due to much noise injected by random memories. When contrastive factor $\alpha$ is set to $0$, it is essentially greedy retrieval, and an important observation is that the translation quality of our model increases with the $\alpha$ until it drops at some certain point. We suspect that too large $\alpha$ would yield mutually contrastive TM that divert too much from the original source sentence. Similar phenomenon can be verified in the Figure \ref{fig:tm_size}, when TM size equals to $0$, it is a non-TM translation model delivering worst result while too large TM size also hurts the model performance which is also observed in ~\citet{fuzzy,xia}.

To further demonstrate the intuition behind our framework, we randomly sample 1,000 examples from test sets of En$\rightarrow$De and En$\rightarrow$Es directions and use t-SNE~\citep{van2008visualizing} to visualize the sentence embedding of translation memories and target sentence encoded by our CMM. The result is shown in Figure \ref{fig:visualization} and one interesting observation is that although the target side of testset is never exposed to the model, the representation of translation memories are uniformly distributed around the target sentence in the latent semantic space. \\
\section{Conclusion}
\label{sec:conclusion}
In this work, we introduce an approach to incorporate contrastive translation memories into a NMT system. Our system demonstrates its superiority in retrieval, memory encoding and training phases. 

Experimental results on four translation datasets verify the effectiveness of our framework. In the future, we plan to exploit the potential of this general idea in different retrieval-generation tasks.

\section{Limitations}
\label{sec:limitations}
This paper propose a framework for Retrieval-augmented Neural Machine Translation and it relies on holistically similar but mutually contrastive translation memories which makes it work mostly for corpora in the same domain. How to apply this general idea to other scenario like low resource NMT remains a future challenge.
\section{Acknowledgement} This work was supported by National Natural Science Foundation of China (NSFC Grant No. 62122089 and No. 61876196). Rui Yan is supported by Tencent Collaborative Research Fund.
\bibliography{emnlp2022}

\begin{thebibliography}{35}
\expandafter\ifx\csname natexlab\endcsname\relax\def\natexlab#1{#1}\fi

\bibitem[{Bult{\'e} and Tezcan(2019)}]{fuzzy}
Bram Bult{\'e} and Arda Tezcan. 2019.
\newblock Neural fuzzy repair: Integrating fuzzy matches into neural machine
  translation.
\newblock In \emph{57th Annual Meeting of the
  Association-for-Computational-Linguistics (ACL)}, pages 1800--1809.

\bibitem[{Cai et~al.(2021)Cai, Wang, Li, Lam, and Liu}]{cai}
Deng Cai, Yan Wang, Huayang Li, Wai Lam, and Lemao Liu. 2021.
\newblock Neural machine translation with monolingual translation memory.
\newblock In \emph{Proceedings of the 59th Annual Meeting of the Association
  for Computational Linguistics and the 11th International Joint Conference on
  Natural Language Processing (Volume 1: Long Papers)}, pages 7307--7318.

\bibitem[{Callison-Burch et~al.(2006)Callison-Burch, Osborne, and
  Koehn}]{callison-burch-etal-2006-evaluating}
Chris Callison-Burch, Miles Osborne, and Philipp Koehn. 2006.
\newblock \href {https://aclanthology.org/E06-1032} {Re-evaluating the role of
  {B}leu in machine translation research}.
\newblock In \emph{11th Conference of the {E}uropean Chapter of the Association
  for Computational Linguistics}, pages 249--256, Trento, Italy. Association
  for Computational Linguistics.

\bibitem[{Cao and Xiong(2018)}]{gated}
Qian Cao and Deyi Xiong. 2018.
\newblock Encoding gated translation memory into neural machine translation.
\newblock In \emph{Proceedings of the 2018 Conference on Empirical Methods in
  Natural Language Processing}, pages 3042--3047.

\bibitem[{Carbonell and Goldstein(1998)}]{mmr}
Jaime Carbonell and Jade Goldstein. 1998.
\newblock The use of mmr, diversity-based reranking for reordering documents
  and producing summaries.
\newblock In \emph{Proceedings of the 21st annual international ACM SIGIR
  conference on Research and development in information retrieval}, pages
  335--336.

\bibitem[{Chen et~al.(2020)Chen, Kornblith, Norouzi, and
  Hinton}]{chen2020simple}
Ting Chen, Simon Kornblith, Mohammad Norouzi, and Geoffrey Hinton. 2020.
\newblock A simple framework for contrastive learning of visual
  representations.
\newblock In \emph{International conference on machine learning}, pages
  1597--1607. PMLR.

\bibitem[{Christensen and Schjoldager(2010)}]{christensen2010translation}
Tina~Paulsen Christensen and Anne Schjoldager. 2010.
\newblock Translation-memory (tm) research: what do we know and how do we know
  it?
\newblock \emph{Hermes-Journal of Language and Communication in Business},
  (44):89--101.

\bibitem[{Feng et~al.(2017)Feng, Zhang, Zhang, Wang, and Abel}]{feng2017memory}
Yang Feng, Shiyue Zhang, Andi Zhang, Dong Wang, and Andrew Abel. 2017.
\newblock Memory-augmented neural machine translation.
\newblock In \emph{Proceedings of the 2017 Conference on Empirical Methods in
  Natural Language Processing}, pages 1390--1399.

\bibitem[{Gao et~al.(2021)Gao, Yao, and Chen}]{simcse}
Tianyu Gao, Xingcheng Yao, and Danqi Chen. 2021.
\newblock Simcse: Simple contrastive learning of sentence embeddings.
\newblock In \emph{Proceedings of the 2021 Conference on Empirical Methods in
  Natural Language Processing}, pages 6894--6910.

\bibitem[{Gilmer et~al.(2017)Gilmer, Schoenholz, Riley, Vinyals, and
  Dahl}]{gilmer2017neural}
Justin Gilmer, Samuel~S Schoenholz, Patrick~F Riley, Oriol Vinyals, and
  George~E Dahl. 2017.
\newblock Neural message passing for quantum chemistry.
\newblock In \emph{International conference on machine learning}, pages
  1263--1272. PMLR.

\bibitem[{Gu et~al.(2016)Gu, Lu, Li, and Li}]{gu2016incorporating}
Jiatao Gu, Zhengdong Lu, Hang Li, and Victor~OK Li. 2016.
\newblock Incorporating copying mechanism in sequence-to-sequence learning.
\newblock In \emph{Proceedings of the 54th Annual Meeting of the Association
  for Computational Linguistics (Volume 1: Long Papers)}, pages 1631--1640.

\bibitem[{Gu et~al.(2018)Gu, Wang, Cho, and Li}]{gu}
Jiatao Gu, Yong Wang, Kyunghyun Cho, and Victor~OK Li. 2018.
\newblock Search engine guided neural machine translation.
\newblock In \emph{Proceedings of the AAAI Conference on Artificial
  Intelligence}, volume~32.

\bibitem[{Hadsell et~al.(2006)Hadsell, Chopra, and
  LeCun}]{hadsell2006dimensionality}
Raia Hadsell, Sumit Chopra, and Yann LeCun. 2006.
\newblock Dimensionality reduction by learning an invariant mapping.
\newblock In \emph{2006 IEEE Computer Society Conference on Computer Vision and
  Pattern Recognition (CVPR'06)}, volume~2, pages 1735--1742. IEEE.

\bibitem[{He et~al.(2020)He, Fan, Wu, Xie, and Girshick}]{moco}
Kaiming He, Haoqi Fan, Yuxin Wu, Saining Xie, and Ross Girshick. 2020.
\newblock Momentum contrast for unsupervised visual representation learning.
\newblock In \emph{Proceedings of the IEEE/CVF Conference on Computer Vision
  and Pattern Recognition}, pages 9729--9738.

\bibitem[{He et~al.(2021)He, Huang, Cui, Li, and Liu}]{qiuxiang}
Qiuxiang He, Guoping Huang, Qu~Cui, Li~Li, and Lemao Liu. 2021.
\newblock Fast and accurate neural machine translation with translation memory.
\newblock In \emph{Proceedings of the 59th Annual Meeting of the Association
  for Computational Linguistics and the 11th International Joint Conference on
  Natural Language Processing (Volume 1: Long Papers)}, pages 3170--3180.

\bibitem[{Khandelwal et~al.(2020)Khandelwal, Fan, Jurafsky, Zettlemoyer, and
  Lewis}]{knn_mt}
Urvashi Khandelwal, Angela Fan, Dan Jurafsky, Luke Zettlemoyer, and Mike Lewis.
  2020.
\newblock Nearest neighbor machine translation.
\newblock In \emph{International Conference on Learning Representations}.

\bibitem[{Lee et~al.(2020)Lee, Lee, and Hwang}]{lee2020contrastive}
Seanie Lee, Dong~Bok Lee, and Sung~Ju Hwang. 2020.
\newblock Contrastive learning with adversarial perturbations for conditional
  text generation.
\newblock In \emph{International Conference on Learning Representations}.

\bibitem[{Liu et~al.(2012)Liu, Cao, Watanabe, Zhao, Yu, and
  Zhu}]{liu-etal-2012-locally}
Lemao Liu, Hailong Cao, Taro Watanabe, Tiejun Zhao, Mo~Yu, and Conghui Zhu.
  2012.
\newblock \href {https://aclanthology.org/D12-1037} {Locally training the
  log-linear model for {SMT}}.
\newblock In \emph{Proceedings of the 2012 Joint Conference on Empirical
  Methods in Natural Language Processing and Computational Natural Language
  Learning}, pages 402--411, Jeju Island, Korea. Association for Computational
  Linguistics.

\bibitem[{Liu et~al.(2019)Liu, Wang, Zong, and Su}]{liu2019unified}
Yang Liu, Kun Wang, Chengqing Zong, and Keh-Yih Su. 2019.
\newblock A unified framework and models for integrating translation memory
  into phrase-based statistical machine translation.
\newblock \emph{Computer Speech \& Language}, 54:176--206.

\bibitem[{Liu and Liu(2021)}]{simcls}
Yixin Liu and Pengfei Liu. 2021.
\newblock Simcls: A simple framework for contrastive learning of abstractive
  summarization.
\newblock In \emph{Proceedings of the 59th Annual Meeting of the Association
  for Computational Linguistics and the 11th International Joint Conference on
  Natural Language Processing (Volume 2: Short Papers)}, pages 1065--1072.

\bibitem[{Mikolov et~al.(2013)Mikolov, Sutskever, Chen, Corrado, and
  Dean}]{mikolov2013distributed}
Tomas Mikolov, Ilya Sutskever, Kai Chen, Greg~S Corrado, and Jeff Dean. 2013.
\newblock Distributed representations of words and phrases and their
  compositionality.
\newblock In \emph{Advances in neural information processing systems}, pages
  3111--3119.

\bibitem[{Pan et~al.(2021)Pan, Wang, Wu, and Li}]{mrasp2}
Xiao Pan, Mingxuan Wang, Liwei Wu, and Lei Li. 2021.
\newblock Contrastive learning for many-to-many multilingual neural machine
  translation.
\newblock In \emph{Proceedings of the 59th Annual Meeting of the Association
  for Computational Linguistics and the 11th International Joint Conference on
  Natural Language Processing (Volume 1: Long Papers)}, pages 244--258.

\bibitem[{See et~al.(2017)See, Liu, and Manning}]{pointer_generator}
Abigail See, Peter~J Liu, and Christopher~D Manning. 2017.
\newblock Get to the point: Summarization with pointer-generator networks.
\newblock In \emph{Proceedings of the 55th Annual Meeting of the Association
  for Computational Linguistics (Volume 1: Long Papers)}, pages 1073--1083.

\bibitem[{Sener and Koltun(2018)}]{sener2018multi}
Ozan Sener and Vladlen Koltun. 2018.
\newblock Multi-task learning as multi-objective optimization.
\newblock \emph{Advances in neural information processing systems}, 31.

\bibitem[{Sennrich et~al.(2016)Sennrich, Haddow, and Birch}]{bpe}
Rico Sennrich, Barry Haddow, and Alexandra Birch. 2016.
\newblock Neural machine translation of rare words with subword units.
\newblock In \emph{54th Annual Meeting of the Association for Computational
  Linguistics}, pages 1715--1725. Association for Computational Linguistics
  (ACL).

\bibitem[{Simard and Isabelle(2009)}]{simard2009phrase}
Michel Simard and Pierre Isabelle. 2009.
\newblock Phrase-based machine translation in a computer-assisted translation
  environment.
\newblock \emph{Proceedings of the Twelfth Machine Translation Summit (MT
  Summit XII)}, pages 120--127.

\bibitem[{Steinberger et~al.(2006)Steinberger, Pouliquen, Widiger, Ignat,
  Erjavec, Tufis, and Varga}]{steinberger2006jrc}
Ralf Steinberger, Bruno Pouliquen, Anna Widiger, Camelia Ignat, Tomaz Erjavec,
  Dan Tufis, and D{\'a}niel Varga. 2006.
\newblock The jrc-acquis: A multilingual aligned parallel corpus with 20+
  languages.
\newblock \emph{arXiv preprint cs/0609058}.

\bibitem[{Sutskever et~al.(2014)Sutskever, Vinyals, and Le}]{seq2seq}
Ilya Sutskever, Oriol Vinyals, and Quoc~V Le. 2014.
\newblock Sequence to sequence learning with neural networks.
\newblock In \emph{Advances in neural information processing systems}, pages
  3104--3112.

\bibitem[{Utiyama et~al.(2011)Utiyama, Neubig, Onishi, and
  Sumita}]{utiyama2011searching}
Masao Utiyama, Graham Neubig, Takashi Onishi, and Eiichiro Sumita. 2011.
\newblock Searching translation memories for paraphrases.
\newblock In \emph{Machine Translation Summit}, volume~13, pages 325--331.

\bibitem[{Van~der Maaten and Hinton(2008)}]{van2008visualizing}
Laurens Van~der Maaten and Geoffrey Hinton. 2008.
\newblock Visualizing data using t-sne.
\newblock \emph{Journal of machine learning research}, 9(11).

\bibitem[{Vaswani et~al.(2017)Vaswani, Shazeer, Parmar, Uszkoreit, Jones,
  Gomez, Kaiser, and Polosukhin}]{transformer}
Ashish Vaswani, Noam Shazeer, Niki Parmar, Jakob Uszkoreit, Llion Jones,
  Aidan~N Gomez, {\L}ukasz Kaiser, and Illia Polosukhin. 2017.
\newblock Attention is all you need.
\newblock In \emph{Advances in neural information processing systems}, pages
  5998--6008.

\bibitem[{Xia et~al.(2019)Xia, Huang, Liu, and Shi}]{xia}
Mengzhou Xia, Guoping Huang, Lemao Liu, and Shuming Shi. 2019.
\newblock Graph based translation memory for neural machine translation.
\newblock In \emph{Proceedings of the AAAI Conference on Artificial
  Intelligence}, volume~33, pages 7297--7304.

\bibitem[{Xu et~al.(2020)Xu, Crego, and Senellart}]{boosting}
Jitao Xu, Josep~M Crego, and Jean Senellart. 2020.
\newblock Boosting neural machine translation with similar translations.
\newblock In \emph{Proceedings of the 58th Annual Meeting of the Association
  for Computational Linguistics}, pages 1580--1590.

\bibitem[{Yamada(2011)}]{yamada2011effect}
Masaru Yamada. 2011.
\newblock The effect of translation memory databases on productivity.
\newblock \emph{Translation research projects}, 3:63--73.

\bibitem[{Zhang et~al.(2018)Zhang, Utiyama, Sumita, Neubig, and
  Nakamura}]{zhang}
Jingyi Zhang, Masao Utiyama, Eiichro Sumita, Graham Neubig, and Satoshi
  Nakamura. 2018.
\newblock Guiding neural machine translation with retrieved translation pieces.
\newblock In \emph{Proceedings of NAACL-HLT}, pages 1325--1335.

\end{thebibliography}
\bibliographystyle{acl_natbib}
\end{document}